%% file: main.tex
\documentclass{article}
\usepackage{spconf,amsmath,bm,graphicx}
\usepackage[dvipsnames]{xcolor}
\usepackage{xcolor}   
\usepackage{tikz}

\usetikzlibrary{positioning}
\usetikzlibrary{shapes.symbols}
\usetikzlibrary{shapes}
\usepackage[symbols,toc,automake,acronym]{glossaries-extra}

\makeglossaries

\usepackage{fontawesome5}

\usetikzlibrary{shadows}

\usepackage{algorithm}
\usepackage{algorithmic}
\usepackage{subcaption}
\usepackage[export]{adjustbox}

\usepackage[pdftex]{hyperref}
\usepackage{amsfonts,amssymb,amsbsy}

\definecolor{pinegreen}{cmyk}{0.92,0,0.59,0.25}
\definecolor{royalblue}{cmyk}{1,0.50,0,0}
\definecolor{lavander}{cmyk}{0,0.48,0,0}
\definecolor{violet}{cmyk}{0.79,0.88,0,0}

\tikzstyle{ncyan}=[circle, draw=cyan!70, thin, fill=white, scale=0.8, font=\fontsize{11}{0}\selectfont]
\tikzstyle{ngreen}=[circle,  draw=green!70, thin, fill=white, scale=0.8, font=\fontsize{11}{0}\selectfont]
\tikzstyle{nred}=[circle, draw=red!70, thin, fill=white, scale=0.8, font=\fontsize{11}{0}\selectfont]
\tikzstyle{ngray}=[circle, draw=gray!70, thin, fill=white, scale=0.55, font=\fontsize{14}{0}\selectfont]
\tikzstyle{nyellow}=[circle, draw=yellow!70, thin, fill=white, scale=0.55, font=\fontsize{14}{0}\selectfont]
\tikzstyle{norange}=[circle,  draw=orange!70, thin, fill=white, scale=0.55, font=\fontsize{10}{0}\selectfont]
\tikzstyle{npurple}=[circle,draw=purple!70, thin, fill=white, scale=0.55, font=\fontsize{10}{0}\selectfont]
\tikzstyle{nblue}=[circle, draw=blue!70, thin, fill=white, scale=0.55, font=\fontsize{10}{0}\selectfont]
\tikzstyle{nteal}=[circle,draw=teal!70, thin, fill=white, scale=0.55, font=\fontsize{10}{0}\selectfont]
\tikzstyle{nviolet}=[circle, draw=violet!70, thin, fill=white, scale=0.55, font=\fontsize{10}{0}\selectfont]
\tikzstyle{qgre}=[rectangle, draw, thin,fill=green!20, scale=0.8]
\tikzstyle{rpath}=[ultra thick, red, opacity=0.4]
\tikzstyle{legend_isps}=[rectangle, rounded corners, thin,fill=gray!20, text=blue, draw]

\title{Comparing Federated Stochastic Gradient Descent and Federated Averaging for Predicting Hospital Length of Stay}
\name{Mehmet Yigit Balik}
\address{Aalto University, Espoo, Finland}
\input{./macros.tex}

\begin{document}
%
\maketitle

\begin{abstract}
Predicting hospital length of stay (LOS) reliably is an essential need for efficient resource allocation at hospitals. Traditional predictive modeling tools frequently have difficulty acquiring sufficient and diverse data because healthcare institutions have privacy rules in place. In our study, we modeled this problem as an empirical graph where nodes are the hospitals. This modeling approach facilitates collaborative model training by modeling decentralized data sources from different hospitals without extracting sensitive data outside of hospitals. A local model is trained on a node (hospital) by aiming the generalized total variation minimization (GTVMin). Moreover, we implemented and compared two different federated learning optimization algorithms named federated stochastic gradient descent (FedSGD) and federated averaging (FedAVG). Our results show that federated learning enables accurate prediction of hospital LOS while addressing privacy concerns without extracting data outside healthcare institutions.
\end{abstract}
\begin{keywords}
Federated Learning, Networks, Personalized Machine Learning, Length of Stay 
\end{keywords}
\section{Introduction}
\label{sec:intro}

In hospitals, efficiently managing resources and managing patient flow are increasingly complicated tasks. Managing how long patients stay in the hospital is seen as a highly effective strategy for smartly handling hospital resources during times of crisis \cite{wang2020survival}. Predicting the length of stay (LOS) of a patient is crucial for meeting the rising demand for healthcare services by making the best use of available resources \cite{ellahham2019use}. The LOS indicates the time frame that time spent by a patient starting from admission to discharge \cite{khosravizadeh2016factors}. 
\\
\\
The LOS is a complicated factor influenced by many aspects, including the patient's medical condition and social circumstances \cite{shea1995computer}, which makes it hard to predict by humans. Research exploring the use of machine learning (ML) methods for LOS in hospitals has become more prevalent in the literature. Many researchers studied predicting LOS by applying centralized ML approaches, where all data is stored in a single environment. In their study, Alsinglawi et al. \cite{alsinglawi2020predicting} constructed a predictive framework utilizing ML regression models to forecast LOS for patients hospitalized with cardiovascular conditions in the Intensive Care Unit. Mekhaldi et al. \cite{mekhaldi2020using} applied ML methods to predict LOS in hospitals. Their methodology includes data preprocessing, model validation, and comparison of outcomes using two different ML  regression models. Turgeman et al. \cite{turgeman2017insights} utilized administrative hospital data to develop a tree-based model aimed at predicting LOS for patients admitted with congestive heart failure upon admission. Tsai et al. \cite{tsai2016length} conducted a comparative analysis between artificial neural networks and linear regression to assess their performance in predicting the length of stay for heart failure inpatients. However, while ML techniques show promise, they often require large amounts of data. Given the limited availability of medical datasets and their sensitive nature, achieving optimal performance can be challenging. Furthermore, conducting the training process using all data simultaneously proves to be both expensive and ineffective. Additionally, consolidating data from all hospitals overlooks the possibility that certain hospitals may exhibit similar patterns, while others may not. In this context, federated learning presents an alternative. Federated learning can improve model accuracy while preserving data privacy by training ML models on decentralized data from various hospitals.
\\
\\
In this study, we structured the problem as an empirical graph where each hospital is a node on the graph. This method allows different hospitals to work together on training local models without needing to share sensitive patient data. We trained a model for each hospital using a method called generalized total variation minimization (GTVMin). Additionally, we tested two different ways of model training, called federated stochastic gradient descent (FedSGD) and federated averaging (FedAVG), and compared their performances.
\\
\\
This paper is outlined as follows: Section \ref{sec:pf} explains the dataset used, empirical graph construction, and formulation of the GTVMin instance for FedSGD and FedAVG. Section \ref{sec_methods} discusses data preparation, feature selection, local model choice, and implemented federated learning algorithms. Section \ref{sec_results} showcases the performances of applied algorithms. Finally, our key findings are summarized in the final section.

\section{Problem Formulation}
\label{sec:pf}
\subsection{Dataset Definition}
In our study, we used \textit{Microsoft Predicting Length of Stay Dataset} \cite{los-dataset}. The dataset contains 100,000 data points representing patients' information (e.g. gender, glucose level, body-mass index, pulse, etc.) and patients' LOS. Table \ref{tab:dataset-details} shows the details of the used data fields in this study. Section \ref{sec_methods} will discuss the applied prepossessing techniques on the features.
A local dataset $\mathcal{D}^{(i)}$ defined as follows:
\begin{equation}\label{local-dataset}
    \mathcal{D}^{(i)} = \{\{\mathbf{x}^{(i, 1)}, y^{(i, 1)}\}, ..., \{\mathbf{x}^{(i, m_i)}, y^{(i, m_i)}\}\}
\end{equation}
Where $\mathbf{x}^{(i, r)}$ and $y^{(i, r)}$ indicate features and the label of the $r^{th}$ data point in the local dataset $\mathcal{D}^{(i)}$. Finally, $m_i$ indicates the size of the local dataset $\mathcal{D}^{(i)}$, which might differ between different nodes.
We can also represent features as a feature matrix and labels as a label vector as follows:
\begin{equation}
    \mathbf{X}^{(i)} = (\mathbf{x}^{(i, 1)}, ..., \mathbf{x}^{(i, m_i)})^T \text{, and } \mathbf{y}^{(i)}  = (y^{(i, 1)}, ..., y^{(i, m_i)})^T
\end{equation}

\begin{table}[htbp]
    \footnotesize
    \centering
    \begin{tabular}{|l|l|p{3cm}|}
        \hline
        \textbf{Data Field} & \textbf{Type} & \textbf{Description} \\ 
        \hline
        \textit{rcount} & Categorical	& Number of readmissions within last 180 days \\
        \hline
        \textit{gender} & String & Gender of the patient - M or F \\ 
        \hline
        \textit{hemo} & String & Flag for blood disorder during encounter \\ 
        \hline
        \textit{hematocrit} & Float & Average hematocrit value during encounter (g/dL) \\ 
        \hline
        \textit{neutrophils} & Float & Average neutrophils value during encounter (cells/microL) \\ 
        \hline
        \textit{sodium} & Float & Average sodium value during encounter (mmol/L) \\ 
        \hline
        \textit{glucose} & Float & Average glucose value during encounter (mmol/L) \\ 
        \hline
        \textit{bloodureanitro} & Float & Average blood urea nitrogen value during encounter (mg/dL) \\ 
        \hline
        \textit{creatinine} & Float & Average creatinine value during encounter (mg/dL) \\ 
        \hline
        \textit{bmi} & Float & Average BMI during encounter (kg/m\textsuperscript{2}) \\ 
        \hline
        \textit{pulse} & Float & Average pulse during encounter (beats/m) \\ 
        \hline
        \textit{respiration} & Float & Average respiration during encounter (breaths/m) \\ 
        \hline
        \textit{n\_conditions} & Integer & Number of clinical conditions that the patient has \\ 
        \hline
        \textit{lengthofstay} (label) & Integer & Length of stay for the encounter \\ 
        \hline
        \textit{facid} (node ID) & Integer & Facility ID at which the encounter occurred \\ 
        \hline
    \end{tabular}
    \caption{Description of Data Fields}
    \label{tab:dataset-details}
\end{table}

\subsection{Empirical Graph}
The empirical graph $\mathcal{G} = \mathcal{(V, E)}$ is an undirected weighted graph with nodes $\mathcal{V} = \{1, ..., n\}$ where each node has a local dataset formulated as (\ref{local-dataset}). Each node has a local model parameterized by $\mathbf{w}^{(i)}$. An undirected edge $(i, j) \in \mathcal{E}$ denotes the similarity between local datasets $\mathcal{D}^{(i)}$ and $\mathcal{D}^{(j)}$ and has the weight $A_{i, j} \in \{0, 1\}$.
\\
\\
We measure the discrepancy between two local datasets $\mathcal{D}^{(i)}$ and $\mathcal{D}^{(j)}$ using a discrepancy measure $D(\mathbf{w}^{(i)}, \mathbf{w}^{(j)})$ (such as Euclidean distance, Kullback–Leibler divergence or Wasserstein distance), which accepts local model weights as its parameters. Given a minimum node degree $d$, we connect this node to other $d$ nodes with the smallest discrepancy and set the corresponding edge weight to 1. During the construction of the empirical graph $\mathcal{G}$, we learn local model parameters $\mathbf{w}^{(i)}$ by minimizing a local loss function $L_i(\mathbf{w}^{(i)})$ on local dataset $\mathcal{D}^{(i)}$ without sharing information between nodes. The motivation behind our approach is to learn weights that represent the local dataset in the best way. Then we can use these weights as the representation of the corresponding local dataset and measure dissimilarity between nodes.

\subsection{Formulation of GTVMin Instance}
Our main objective is to learn the best local model parameters $\hat{\mathbf{w}}^{(i)}$ by minimizing their local loss as well as enforcing the small total variation \cite{jung24fl}. In this paper, we implemented two different federated optimization algorithms.

\subsubsection{FedSGD}
The formulation of GTVMin is given as follows \cite{jung24fl}:
\begin{equation} \label{gtvmin-fedsgd}
\begin{split}
    \textbf{stack} \{\hat{\mathbf{w}}^{(i)}\}_{i = 1}^n \in \argmin_{\textbf{stack} \{\mathbf{w}^{(i)}\}_{i = 1}^n} \sum_{i \in \mathcal{V}}  L_i(\mathbf{w}^{(i)}) \\ + \alpha \sum_{(i, j) \in \mathcal{E}} A_{i, j} \norm{\mathbf{w}^{(i)} - \mathbf{w}^{(j)}}_2^2
\end{split}
\end{equation}

\subsubsection{FedAVG}

We can also achieve same local model weights $\mathbf{\hat{w}}^{(i)}$ and solve GTVMin as follows \cite{jung24fl}:
\begin{equation} \label{gtvmin-fedavg}
\begin{gathered}
    \mathbf{\hat{w}} \in \argmin_{\mathbf{w} \in \mathcal{C}} \sum_{i \in \mathcal{V}}  L_i(\mathbf{w}^{(i)}) \\
    \text{with } \mathcal{C} = \{\mathbf{w} = \textbf{stack} \{\mathbf{w}^{(i)}\}_{i = 1}^n : \mathbf{w}^{(i)} = \mathbf{w}^{(j)} \\ \text{ for any } i, j \in \mathcal{V}\}
\end{gathered}
\end{equation}

\section{Methods} 
\label{sec_methods} 
\subsection{Data Preparation and Feature Selection}\label{data-prep}
Dataset contains data collected from 5 different hospital facilities. Table \ref{loca-dataset-sizes} gives the number of data points in each local dataset. Each local dataset is divided into training, validation, and test sets randomly using seed 42 with percentages of 70\%, 15\%, and 15\%, respectively. We use standard split to ensure that the model is robust, generalizes well to new data, and avoids overfitting, leading to more reliable and accurate machine learning applications. The validation set is used to tune hyperparameters such as $\alpha$ value in (\ref{gtvmin-fedsgd}), the minimum node degree $d$ and the learning rate $\eta$ in (\ref{gradient-fedsgd}), (\ref{gradient-fedavg1}) and (\ref{gradient-fedavg2}).
\begin{table}[H]
    \small
    \centering
    \begin{tabular}{||c|c||} 
     \hline
     \textbf{Local Dataset} & \textbf{Dataset Size} \\
     \hline\hline
     $\mathcal{D}^{(1)}$ & 30012 \\ 
     \hline
     $\mathcal{D}^{(2)}$ & 30035 \\
     \hline
     $\mathcal{D}^{(3)}$ & 30755 \\
     \hline
     $\mathcal{D}^{(4)}$ & 4499 \\
     \hline
     $\mathcal{D}^{(5)}$ & 4699 \\
     \hline
    \end{tabular}
    \caption{Sizes of Local Datasets}
    \label{loca-dataset-sizes}
\end{table}
\noindent
There were 28 data fields including the label and facility ID (i.e. LOS value) in the dataset \cite{los-dataset}. However, irrelevant features (admission id, date of discharge, etc.) are dropped and the remaining features are indicated in Table \ref{tab:dataset-details}. We dropped those features since we aim to infer LOS utilizing the medical information of the patient rather than dates of admission and discharge or ID of the admission. Binary features (\textit{gender} and \textit{hemo}) are used without any preprocessing. There are also different binary features indicating the patient's conditions such as depression and asthma in the original dataset. However, we decided to create a new feature named \textit{n\_conditions} that represents the number of conditions a patient has and we dropped the binary condition features for the sake of simplicity. Categorical feature \textit{rcount} $\in$ \{"0", "1", "2", "3", "4", "5+"\} is converted one-hot-encoding format where we fixed value "5+" to 5 to eliminate ambiguity. The numerical features (\textit{hematocrit}, \textit{neutrophils}, \textit{sodium}, \textit{glucose}, \textit{bloodureanitro}, \textit{creatinine}, \textit{bmi}, \textit{pulse}, and \textit{respiration}) are normalized as follows:
\\
\\
Let $x^{(i, r)}_j$ represent the value of the  $j^{th}$ feature in the $r^{th}$ data point in the local training set of node $i$. Then, the mean and standard deviation of the $j$-th feature in the local training set can be denoted as $\bar{x}_j^{(i)}$ and $\sigma_j^{(i)}$, respectively. The normalization of the $j$-th feature in both the local training and test sets can be expressed as:
\[
\text{For the training set:} \quad x_j^{(i, r)} \leftarrow \frac{x_j^{(i, r)} - \bar{x}_j^{(i)}}{\sigma_j^{(i)}}
\]
\[
\text{For the validation/test set:} \quad x_j^{(i, r')} \leftarrow \frac{x_j^{(i, r')} - \bar{x}_j^{(i)}}{\sigma_j^{(i)}}
\]
where $r'$ denotes the index of a data point in the local validation/test set of node $i$.

\subsection{Local Model}
Local linear models (\ref{linear-model}) are preferred for predicting LOS due to their interpretability, efficiency, and robustness, making them suitable for healthcare analytics. Their transparent interpretations of coefficients allow for understanding the impact of predictor variables on outcomes, crucial in clinical decision-making. Additionally, their computational efficiency and scalability enable handling large datasets and real-time predictions in healthcare settings effectively. 

\begin{equation} \label{linear-model}
    \mathbf{y}^{(i)} \approx \mathbf{X}^{(i)} \mathbf{w}^{(i)}
\end{equation}
We selected Euclidean distance as the variation measure between two local datasets $\mathcal{D}^{(i)}$ and $\mathcal{D}^{(j)}$ by projecting the corresponding learned local model parameters to Euclidean space shown in (\ref{euclidean-dist}). We choose Euclidean distance due to its intuitive interpretation, mathematical simplicity, and compatibility with local linear models, ensuring effective comparison between local models. Its straightforward computation and consideration of both the magnitude and direction of differences make it a suitable and widely accepted metric for assessing dissimilarity between parameter vectors.

\begin{equation}\label{euclidean-dist}
    D(\mathbf{w}^{(i)}, \mathbf{w}^{(j)}) = \norm{\mathbf{w}^{(i)} - \mathbf{w}^{(j)}}_2
\end{equation}
In the context of predicting LOS with local linear models, the mean squared error (MSE) loss function is well-suited for training. MSE penalizes the squared difference between predicted and actual LOS values, providing a smooth optimization landscape conducive to efficient model training. More importantly, it is a good indication that the model prediction deviates from the ground truth. Hence, we also use MSE as the evaluation metric during the testing.

\begin{equation}\label{loss-func}
    L_i(\mathbf{w}^{(i)}) = \frac{1}{m_i} \norm{\mathbf{y}^{(i)} - \mathbf{X}^{(i)} \mathbf{w}^{(i)}}_2^2
\end{equation}

\subsection{Federated Learning Algorithms}
\label{FL-algo}

\subsubsection{FedSGD}

FedSGD employs a gradient descent (GD) approach to solve (\ref{gtvmin-fedsgd}). This algorithm enforces similar weights across nodes by message passing between neighboring nodes where the construction of edges plays an important role \cite{jung24fl}. Variation between nodes penalizes weight.

\begin{equation}\label{gradient-fedsgd}
\begin{split}
    \mathbf{w}^{(i, k + 1)} \leftarrow  \mathbf{w}^{(i, k)} + \eta [- \nabla L_i(\mathbf{w}^{(i, k)}) \\ + 2\alpha \sum_{(i, j) \in \mathcal{E}} A_{i, j} (\mathbf{w}^{(i)} - \mathbf{w}^{(j)})]
\end{split}
\end{equation}
Moreover, FedSGD does not operate on the whole training dataset at a time. It randomly samples $b$ training data from the training set to represent the whole training data.

\subsubsection{FedAVG Version 1 (FedAVGv1)}
FedAVGv1 employs a projected GD approach to solve (\ref{gtvmin-fedavg}) by (i) taking a local gradient step in each node, then (ii) sending updated weights to a center, and finally (iii) calculating the average of updated weights and distributing it to the nodes \cite{jung24fl}. The mathematical formulation is given as:

\begin{equation}\label{gradient-fedavg1}
\begin{gathered}
    \mathbf{\hat{w}}^{(i)}_k \leftarrow \mathbf{w}^{(i, k)} - \eta \nabla  L_i(\mathbf{w}^{(i, k)}) \\
    \text{\footnotesize (local gradient step)}
\end{gathered}
\end{equation}

\begin{equation}
\begin{gathered}
    \mathbf{w}^{(i, k+1)} \leftarrow \frac{1}{n} \sum_{i' \in \mathcal{V}} \mathbf{\hat{w}}^{(i')}_k\\
    \text{\footnotesize (projection)}
\end{gathered}
\end{equation}

\subsubsection{FedAVG Version 2 (FedAVGv2)}
In FedAVGv1, the client takes a gradient step and waits for global weights. In some situations, this waiting time can be long and wasted. Hence, a client can use that time efficiently by using a local minimization around global weights \cite{jung24fl}. FedAVGv2 follows the procedure (i) local minimization around the global weights, (ii) sending updated weights to a center, and finally (iii) calculating the average of updated weights and distributing it to the nodes \cite{jung24fl} to solve (\ref{gtvmin-fedavg}). The mathematical formulation is given as:

\begin{equation}\label{gradient-fedavg2}
\begin{gathered}
    \mathbf{\hat{w}}^{(i)}_k \leftarrow \argmin_{\mathbf{v}} L_i(\mathbf{v}) + (1/\eta) \norm{\mathbf{v} - \mathbf{w}^{(i, k)}}_2^2\\
    \text{\footnotesize(local minimization around global weights)}
\end{gathered}
\end{equation}

\begin{equation}
\begin{gathered}
    \mathbf{w}^{(i, k+1)} \leftarrow \frac{1}{n} \sum_{i' \in \mathcal{V}} \mathbf{\hat{w}}^{(i')}_k\\
    \text{\footnotesize(update of global weights)}
\end{gathered}
\end{equation}
In FedAVG, the edges do not have any effect on the training since this algorithm operates in a server-client paradigm (the empirical graph is assumed to be a star graph where all nodes have only one neighbor which is the server) \cite{jung24fl}. Similar weights are enforced by averaging local weights to obtain global weights and distributing these global weights to nodes again.  

\section{Results} 
\label{sec_results} 
\begin{figure}[H]
    \centering
    \includegraphics [width=0.9\linewidth]{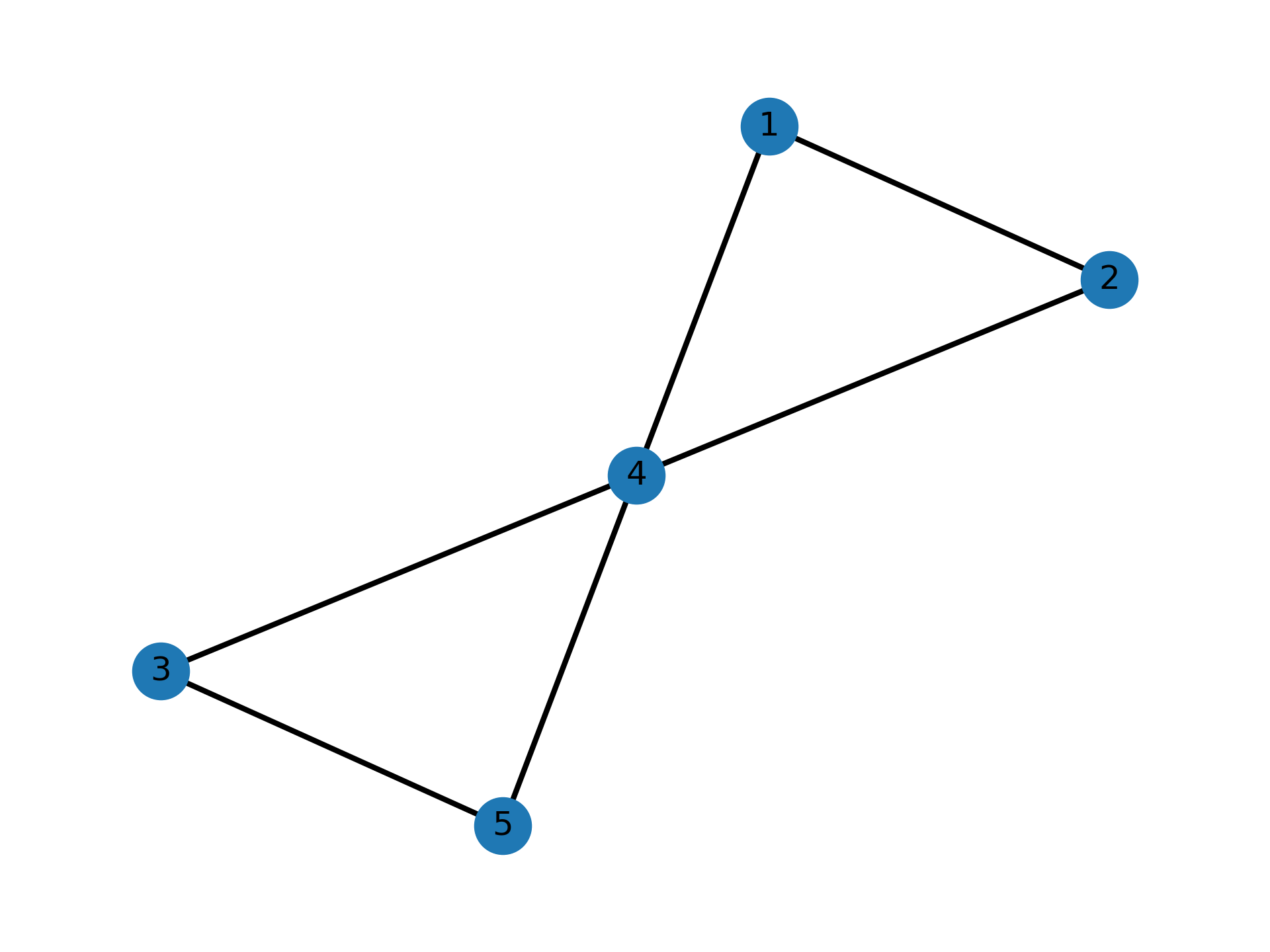}
    \caption{Constructed Empirical Graph Used in FedSGD}
    \label{fedsgd-graph}
\end{figure}

Tuning hyperparameters is a crucial step to obtain the most suitable empirical graph and the optimal local model weights. FedSGD algorithm has 3 hyperparameters listed below:
\begin{itemize}
    \item $\alpha$ value in (\ref{gtvmin-fedsgd}): Determines importance of variation across different model weights.
    \item The minimum node degree $d$: Determines the minimum number of neighbors for a node $i$.
    \item The learning rate $\eta$ in (\ref{gradient-fedsgd}), (\ref{gradient-fedavg1}) and (\ref{gradient-fedavg2}): Determines the size of gradient step.
\end{itemize}
In FedAVG algorithms, we assume a star graph without explicitly constructing the empirical graph. Moreover, the similarity of local model weights is enforced by averaging at a server node without needing $\alpha$. Hence, the only hyperparameter for FedAVG algorithms is $\eta$. We fixed the maximum number of iterations to $1000$ for both algorithms and set batch size $b = 512$ for the FedSGD algorithm. We did not tune these hyperparameters. These hyperparameters were selected among $\alpha \in \{1, 0.5, 0.1\}$, $\eta \in \{0.1, 0.01, 0.001\}$, and $d \in \{1, 2, 3, 4\}$, with the additional constraint that the graph must be connected, based on the lowest error observed on the validation set. For FedSGD, optimal parameters were $\alpha = 0.1$, $\eta = 0.1$ and $d = 2$. The constructed empirical graph is shown in Figure \ref{fedsgd-graph} (recall that the edge weights are 1). For both FedAVG algorithms, the optimal learning rate obtained is $\eta = 0.1$.

\begin{figure}[H]
    \centering
    \includegraphics [width=0.9\linewidth]{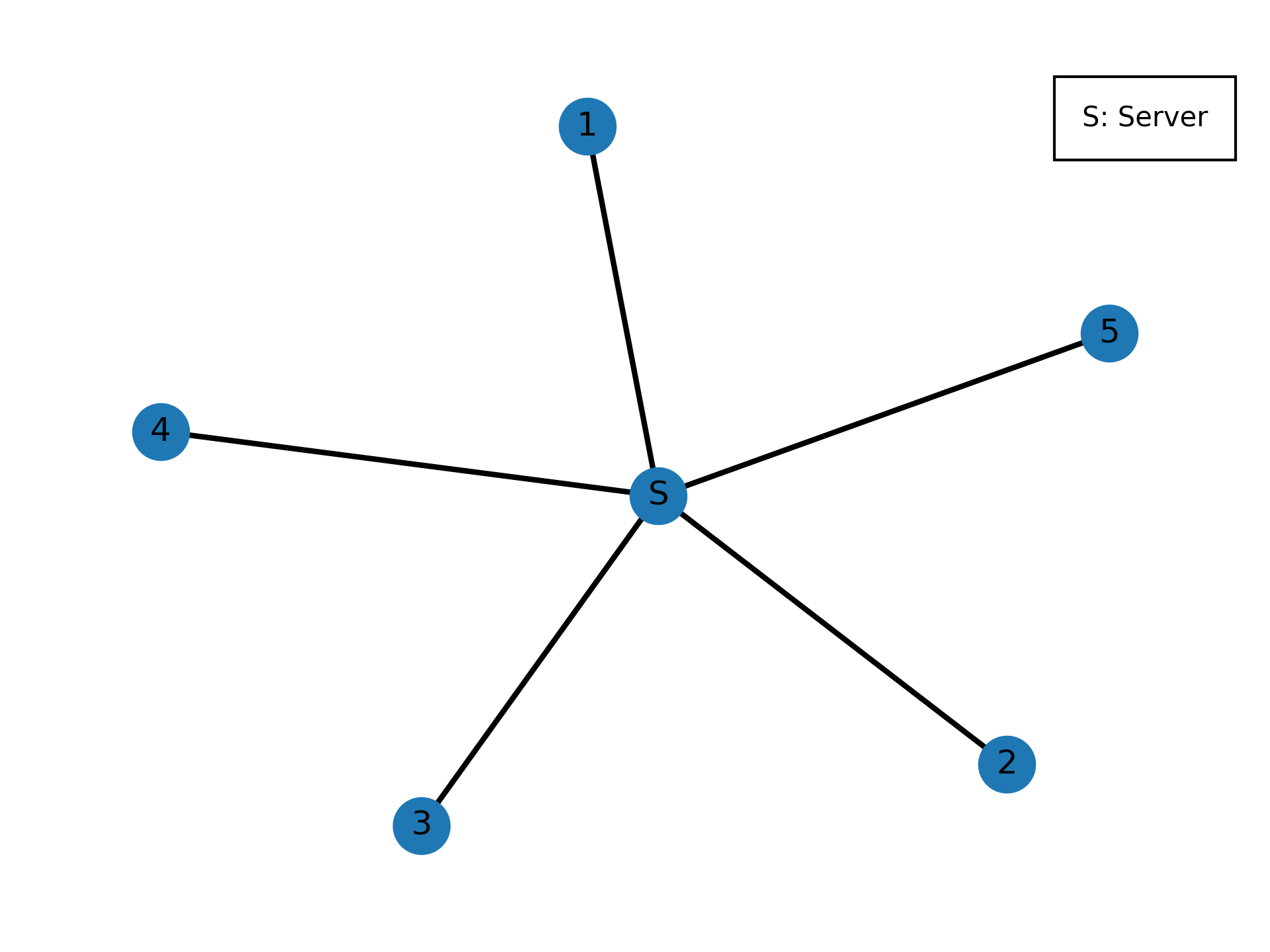}
    \caption{Assumed Empirical Graph in FedAVG}
    \label{fedavg-graph}
\end{figure}

\noindent
Table \ref{train-err} and Table \ref{val-err} depict the training and validation errors obtained with these optimal hyperparameters, respectively. Although all three algorithms have MSE less than 2, the FedSGD algorithm has outperformed FedAVG algorithms, achieving an MSE of 1.377 on the training set and 1.407 on the validation set, which is approximately 0.4 less error compared to other algorithms. Hence, FedSGD is chosen as more applicable in our study. 
\begin{table}[H]
\centering
\begin{tabular}{|c|c|c|c|}
\hline
&\textbf{FedSGD}&\textbf{FedAVGv1}&\textbf{FedAVGv2}\\
\hline 
\textbf{Node} & \begin{tabular}{@{}c@{}}\textbf{\textit{Training}} \\ \textbf{\textit{MSE}} \end{tabular} & \begin{tabular}{@{}c@{}}\textbf{\textit{Training}} \\ \textbf{\textit{MSE}} \end{tabular} & \begin{tabular}{@{}c@{}}\textbf{\textit{Training}} \\ \textbf{\textit{MSE}} \end{tabular}\\
\hline
1 & 1.136 & 1.728 & 1.828\\
2 & 1.123 & 1.724 & 1.829\\
3 & 1.603 & 2.104 & 2.236\\
4 & 1.52  & 1.704 & 1.767\\
5 & 1.503 & 1.819 & 1.895\\
\hline
mean & \textbf{1.377} & 1.816 & 1.911\\
\hline
\end{tabular}
\caption{Training Errors}\label{train-err}
\end{table}
\noindent
It is important to note that for each algorithm and each node, training, and validation errors are similar, which indicates the training was done properly without any overfitting.
\begin{table}[H]
\centering
\begin{tabular}{|c|c|c|c|}
\hline
&\textbf{FedSGD}&\textbf{FedAVGv1}&\textbf{FedAVGv2}\\
\hline 
\textbf{Node} & \begin{tabular}{@{}c@{}}\textbf{\textit{Validation}} \\ \textbf{\textit{MSE}} \end{tabular} & \begin{tabular}{@{}c@{}}\textbf{\textit{Validation}} \\ \textbf{\textit{MSE}} \end{tabular} & \begin{tabular}{@{}c@{}}\textbf{\textit{Validation}} \\ \textbf{\textit{MSE}} \end{tabular}\\
\hline
1 & 1.142 & 1.749 & 1.843\\
2 & 1.214 & 1.81  & 1.913\\
3 & 1.585 & 2.085 & 2.216\\
4 & 1.577 & 1.764 & 1.839\\
5 & 1.514 & 1.763 & 1.827\\
\hline
mean & \textbf{1.407} & 1.834 & 1.928\\
\hline
\end{tabular}
\caption{Validation Errors}\label{val-err}
\end{table}
\noindent
Table \ref{test-err} shows the achieved test errors (on the test set which is neither used for training nor validation see Section \ref{data-prep}) by FedSGD, FedAVGv1 and FedAVGv2. Again, the FedSGD algorithm has shown way better performance, achieving an MSE of 1.354 compared to FedAVGv1 wıth 1.798 MSE and FedAVGv2 with 1.897 MSE.
\begin{table}[H]
\centering
\begin{tabular}{|c|c|c|c|}
\hline
&\textbf{FedSGD}&\textbf{FedAVGv1}&\textbf{FedAVGv2}\\
\hline 
\textbf{Node} & \textbf{\textit{Test MSE}} & \textbf{\textit{Test MSE}} & \textbf{\textit{Test MSE}}\\
\hline
1 & 1.061 & 1.666 & 1.767\\
2 & 1.097 & 1.706 & 1.81\\
3 & 1.685 & 2.224 & 2.358\\
4 & 1.361 & 1.525 & 1.592\\
5 & 1.565 & 1.87  & 1.957\\

\hline
mean & \textbf{1.354} & 1.798 & 1.897\\
\hline
\end{tabular}
\caption{Test Errors}\label{test-err}
\end{table}

\section{Conclusion}

In this paper, we have modeled the problem of predicting hospital LOS as an empirical graph and trained local linear models in nodes to solve GTVMin. The empirical graph is constructed by measuring the Euclidean distance between nodes using the local model weights. As the local model, we selected linear models and MSE as the loss function. We have implemented and compared three different federated learning algorithms, named FedSGD and two different versions of FedAVG, to predict hospital LOS. In the first version of FedAVG (FedAVGv1), a local model takes a single gradient step over the global weights and sends the updated local weights to the center for federated averaging. On the other hand second version (FedAVGv2) local models employ a local minimization around the global weights and then send the updated local weights to the center for averaging. \\

\noindent
Experimental results show that all three algorithms perform accurately and achieve a satisfactory and competitive MSE values on test sets, comparable to existing methods. However, FedSGD outperformed both FedAVGv1 and FedAVGv2, achieving the lowest MSE on training, validation, and test sets. The main reason for FedSGD's superior performance can be attributed to the fact that facilities do not necessarily have similar data. This allows FedSGD to tune the $\alpha$ parameter, thereby adjusting the strength of information sharing between different facilities. In contrast, FedAVG-based algorithms average local model weights and broadcast them to the nodes, which can have a negative impact when different facilities contain highly dissimilar data. \\

\noindent
Although our study presents promising results with proper training without overfitting (the maximum difference between training error and validation error is around 0.1), there are areas for improvement. Firstly, the choice of hyperparameters, such as the learning rate and the minimum node degree, could be further optimized to enhance model performance. Additionally, investigating alternative discrepancy measures and federated learning algorithms could lead to even better results. Finally, extending the study to include more decentralized datasets could provide insights into the applicability of the proposed approach to real life. We leave these improvements for future work.

\newpage
\bibliographystyle{IEEEtran}
\bibliography{ref}

\end{document}

%% file: macros.tex
\newcommand{\norm}[1]{\Vert  {#1} \Vert}

\newcommand{\bmx}[0]{\begin{bmatrix}}
\newcommand{\emx}[0]{\end{bmatrix}}

\DeclareMathOperator*{\argmin}{argmin}

